# A Bayesian Approach to Learning Bayesian Networks with Local Structure

**David Maxwell Chickering**
dmax@microsoft.com

**David Heckerman**
heckerma@microsoft.com

**Christopher Meek**
meek@microsoft.com

Microsoft Research
Redmond WA, 98052-6399

## Abstract

Recently several researchers have investigated techniques for using data to learn Bayesian networks containing compact representations for the conditional probability distributions (CPDs) stored at each node. The majority of this work has concentrated on using decision-tree representations for the CPDs. In addition, researchers typically apply non-Bayesian (or asymptotically Bayesian) scoring functions such as MDL to evaluate the goodness-of-fit of networks to the data.

In this paper we investigate a Bayesian approach to learning Bayesian networks that contain the more general *decision-graph* representations of the CPDs. First, we describe how to evaluate the posterior probability—that is, the Bayesian score—of such a network, given a database of observed cases. Second, we describe various search spaces that can be used, in conjunction with a scoring function and a search procedure, to identify one or more high-scoring networks. Finally, we present an experimental evaluation of the search spaces, using a greedy algorithm and a Bayesian scoring function.

## 1 INTRODUCTION

Given a set of observations in some domain, a common problem that a data analyst faces is to build one or more models of the process that generated the data. In the last few years, researchers in the UAI community have contributed an enormous body of work to this problem, using Bayesian networks as the model of choice. Recent works include Cooper and Herskovits (1992), Buntine (1991), Spiegelhalter et. al (1993), and Heckerman et al. (1995).

A substantial amount of the early work on learning Bayesian networks has used observed data to infer *global* independence constraints that hold in the domain of interest. Global independences are precisely those that follow from the missing edges within a Bayesian-network structure. More recently, researchers (including Boutilier et al., 1995 and Friedman and Goldszmidt, 1996) have extended the "classical" definition of a Bayesian network to include efficient representations of *local* constraints that can hold among the parameters stored in the nodes of the network. Two notable features about the this recent work are (1) the majority of effort has concentrated on inferring decision trees, which are structures that can explicitly represent some parameter equality constraints and (2) researchers typically apply non-Bayesian (or asymptotically Bayesian) scoring functions such as MDL as to evaluate the goodness-of-fit of networks to the data.

In this paper, we apply a *Bayesian* approach to learning Bayesian networks that contain *decision-graphs*—generalizations of decision trees that can encode arbitrary equality constraints—to represent the conditional probability distributions in the nodes.

In Section 2, we introduce notation and previous relevant work. In Section 3 we describe how to evaluate the Bayesian score of a Bayesian network that contains decision graphs. In Section 4, we investigate how a search algorithm can be used, in conjunction with a scoring function, to identify these networks from data. In Section 5, we use data from various domains to evaluate the learning accuracy of a greedy search algorithm applied the search spaces defined in Section 4. Finally, in Section 6, we conclude with a discussion of future extensions to this work.

## 2 BACKGROUND

In this section, we describe our notation and discuss previous relevant work. Throughout the remainder of this paper, we use lower-case letters to refer to variables, and upper-case letters to refer to sets of variables. We write $x_i = k$ when we observe that variable $x_i$ is in state $k$. When we observe the state of every variable in a set $X$, we call the set of observations a *state* of $X$. Although arguably an abuse of notation, we find it convenient to index the states of a set of variables with a single integer. For example, if $X = \{x_1, x_2\}$ is a set containing two binary variables, we may write $X = 2$ to denote $\{x_1 = 1, x_2 = 0\}$.

In Section 2.1, we define a Bayesian network. In Section 2.2 we describe decision trees and how they can be used to represent the probabilities within a Bayesian network. In Section 2.3, we describe decision graphs, which are generalizations of decision trees.

### 2.1 BAYESIAN NETWORKS

Consider a domain $U$ of $n$ discrete variables $x_1, \ldots, x_n$, where each $x_i$ has a finite number of states. A Bayesian network for $U$ represents a joint probability distribution over $U$ by encoding (1) assertions of conditional independence and (2) a collection of probability distributions. Specifically, a Bayesian network $B$ is the pair $(B_S, \Theta)$, where $B_S$ is the *structure* of the network, and $\Theta$ is a set of parameters that encode local probability distributions.

The structure $B_S$ has two components: the *global structure* $\mathcal{G}$ and a set of *local structures* $M$. $\mathcal{G}$ is an acyclic, directed graph—*dag* for short—that contains a node for each variable $x_i \in U$. The edges in $\mathcal{G}$ denote probabilistic dependences among the variables in $U$. We use $Par(x_i)$ to denote the set of parent nodes of $x_i$ in $\mathcal{G}$. We use $x_i$ to refer to both the variable in $U$ and the corresponding node in $\mathcal{G}$. The set of local structures $M = \{M_1, \ldots, M_n\}$ is a set of $n$ mappings, one for each variable $x_i$, such that $M_i$ maps each value of $\{x_i, Par(x_i)\}$ to a parameter in $\Theta$.

The assertions of conditional independence implied by the global structure $\mathcal{G}$ in a Bayesian network $B$ impose the following decomposition of the joint probability distribution over $U$:

$$p(U|B) = \prod_i p(x_i|Par(x_i), \Theta, M_i, \mathcal{G}) \qquad (1)$$

The set of parameters $\Theta$ contains—for each node $x_i$, for each state $k$ of $x_i$, and for each parent state $j$—a single parameter[1] $\Theta(i, j, k)$ that encodes the condi-tional probabilities given in Equation 1. That is,

$$p(x_i = k|Par(x_i) = j, \Theta, M_i, \mathcal{G}) = \Theta(i, j, k) \qquad (2)$$

[1]Because the sum $\sum_k p(x_i = k|Par(x_i), \Theta, M_i, \mathcal{G})$ must

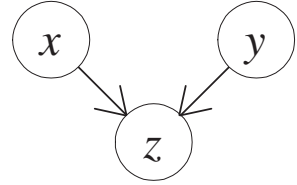

Figure 1: Bayesian network for $U = \{x, y, z\}$

Note that the function $\Theta(i, j, k)$ depends on both $M_i$ and $\mathcal{G}$. For notational simplicity we leave this dependency implicit.

Let $r_i$ denote the number of states of variable $x_i$, and let $q_i$ denote the number of states of the set $Par(x_i)$. We use $\Theta_{ij}$ to denote the set of parameters characterizing the distribution $p(x_i|Par(x_i) = j, \Theta, M_i, \mathcal{G})$:

$$\Theta_{ij} = \cup_{k=1}^{r_i} \Theta(i, j, k)$$

We use $\Theta_i$ to denote the set of parameters characterizing all of the conditional distributions $p(x_i|Par(x_i), \Theta, M_i, \mathcal{G})$:

$$\Theta_i = \cup_{j=1}^{q_i} \Theta_{ij}$$

In the "classical" implementation of a Bayesian network, each node $x_i$ stores $(r_i - 1) \cdot q_i$ distinct parameters in a large table. That is, $M_i$ is simply a lookup into a table. Note that the size of this table grows exponentially with the number of parents $q_i$.

### 2.2 DECISION TREES

There are often equality constraints that hold among the parameters in $\Theta_i$, and researchers have used mappings other than complete tables to more efficiently represent these parameters. For example, consider the global structure $\mathcal{G}$ depicted in Figure 1, and assume that all nodes are binary. Furthermore, assume that if $x = 1$, then the value of $z$ does not depend on $y$. That is,

$$p(z|x = 1, y = 0, \Theta, M_z, \mathcal{G}) = p(z|x = 1, y = 1, \Theta, M_z, \mathcal{G})$$

Using the *decision tree* shown in Figure 2 to implement the mapping $M_z$, we can represent $p(z|x = 1, y, \Theta, M_Z)$ using a single distribution for both $p(z|x = 1, y = 0, \Theta, M_z, \mathcal{G})$ and $p(z|x = 1, y = 1, \Theta, M_z, \mathcal{G})$.

be one, $\Theta$ will actually only contain $r_i - 1$ distinct parameters for this distribution. For simplicity, we leave this implicit for the remainder of the paper.

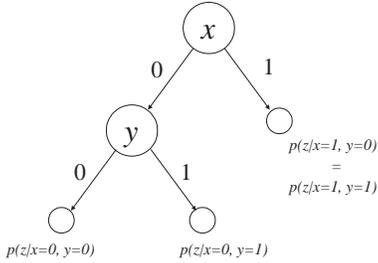

Figure 2: Decision tree for node $z$

Decision trees, described in detail by Breiman (1984), can be used to represent sets of parameters in a Bayesian network. Each tree is a dag containing exactly one root node, and every node other than the root node has exactly one parent. Each leaf node contains a table of $k-1$ distinct parameters that collectively define a conditional probability distribution $p(x_i|Par(x_i), \Theta, M_i, \mathcal{D})$. Each non-leaf node in the tree is annotated with the name of one of the parent variables $\pi \in Par(x_i)$. Out-going edges from a node $\pi$ in the tree are annotated with mutually exclusive and collectively exhaustive sets of values for the variable $\pi$.

When a node $v$ in a decision tree is annotated with the name $\pi$, we say that $v$ *splits* $\pi$. If the edge from $v_1$ to child $v_2$ is annotated with the value $k$, we say that $v_2$ is the child of $v_1$ *corresponding* to $k$. Note that by definition of the edge annotations, the child of a node corresponding to any value is unique.

We traverse the decision tree to find the parameter $\Theta(i,j,k)$ as follows. First, initialize $v$ to be the root node in the decision tree. Then, as long as $v$ is not a leaf, let $\pi$ be the node in $Par(x_i)$ that $v$ splits, and reset $v$ to be the child of $v$ corresponding to the value of $\pi$—determined by $Par(x_i) = j$—and repeat. If $v$ is a leaf, we we return the parameter in the table corresponding to state $k$ of $x_i$.

Decision tree are more expressive mappings than complete tables, as we can represent all of the parameters from a complete table using a *complete decision tree*. A complete decision tree $\mathcal{T}_i$ for a node $x_i$ is a tree of depth $|Par(x_i)|$, such that every node $v_l$ at level $l$ in $\mathcal{T}_i$ splits on the *l*th parent $\pi_l \in Par(x_i)$ and has exactly $r_{\pi_l}$ children, one for each value of $\pi$. It follows by this definition that if $\mathcal{T}_i$ is a complete tree, then $\Theta(i,j,k)$ will map to a distinct parameter for each distinct $\{i,j\}$, which is precisely the behavior of a complete table.

Researchers have found that decision trees are useful for eliciting probability distributions, as experts often have extensive knowledge about equality of conditional distributions. Furthermore, many researchers have developed methods for learning these local structures from data.

### 2.3 DECISION GRAPHS

In this section we describe a generalization of the decision tree, known as a *decision graph*, that can represent a much richer set of equality constraints among the local parameters. A decision graph is identical to a decision tree except that, in a decision graph, the non-root nodes can have more than one parent. Consider, for example, the decision graph depicted in Figure 3. This decision graph represents a conditional probability distribution $p(z|x,y,\Theta)$ for the node $z$ in Figure 1 that has different equality constraints than the tree shown in Figure 2. Specifically, the decision graph encodes the equality

$$p(z|x=0, y=1, \Theta) = p(z|x=1, y=0, \Theta)$$

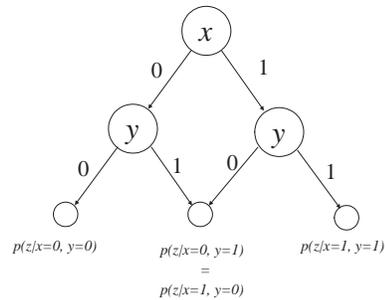

Figure 3: Decision graph for node $z$

We use $\mathcal{D}_i$ to denote a decision graph for node $x_i$. If the mapping in a node $x_i$ is implemented with $\mathcal{D}_i$, we use $\mathcal{D}_i$ instead of $M_i$ to denote the mapping. A decision-graph $\mathcal{D}_i$ can explicitly represent an arbitrary set of equality constraints of the form

$$\Theta_{ij} = \Theta_{ij'} \qquad (3)$$

for $j \neq j'$. To demonstrate this, consider a complete tree $\mathcal{T}_i$ for node $x_i$. We can transform $\mathcal{T}_i$ into a decision graph that represents all of the desired constraints by simply merging together any leaf nodes that contain sets that are equal.

It is interesting to note that any equality constraint of the form given in Equation 3 can also be interpreted as the following *independence* constraint:

$$x_i \perp\!\!\!\perp Par(x_i) \mid Par(x_i) = j \text{ or } Par(x_i) = j'$$

If we allow nodes in a decision graph $\mathcal{D}_i$ to split on node $x_i$ as well as the nodes in $Par(x_i)$, we can represent an arbitrary set of equality constraints among the parameters $\Theta_i$. We return to this issue in Section 6, and assume for now that nodes in $\mathcal{D}_i$ do not split on $x_i$.

## 3 LEARNING DECISION GRAPHS

Many researchers have derived the Bayesian measure-of-fit—herein called the Bayesian *score*—for a network, assuming that there are no equalities among the parameters. Friedman and Goldszmidt (1996) derive the Bayesian score for a structure containing decision trees. In this section, we show how to evaluate the Bayesian score for a structure containing decision graphs.

To derive the Bayesian score, we first need to make an assumption about the process that generated the database $D$. In particular, we assume that the database $D$ is a random (exchangeable) sample from some unknown distribution $\Theta_U$, and that all of the constraints in $\Theta_U$ can be represented using a network structure $B_S$ containing decision graphs.

As we saw in the previous section, the structure $B_S = \{\mathcal{G}, M\}$ imposes a set of independence constraints that must hold in any distribution represented using a Bayesian network with that structure. We define $B_S^h$ to be the *hypothesis* that (1) the independence constraints imposed by structure $B_S$ hold in the joint distribution $\Theta_U$ from which the database $D$ was generated, and (2) $\Theta_U$ contains no other independence constraints. We refer the reader to Heckerman et al. (1994) for a more detailed discussion of structure hypotheses.

The Bayesian score for a structure $B_S$ is the posterior probability of $B_S^h$, given the observed database $D$:

$$p(B_S^h|D) = c \cdot p(D|B_S^h)p(B_S^h)$$

where $c = \frac{1}{p(D)}$. If we are only concerned with the *relative* scores of various structures, as is almost always the case, then the constant $c$ can be ignored. Consequently, we extend our definition of the Bayesian score to be any function proportional to $p(D|B_S^h)p(B_S^h)$.

For now, we assume that there is an efficient method for assessing $p(B_S^h)$ (assuming this distribution is uniform, for example), and concentrate on how to derive the *marginal likelihood* term $p(D|B_S^h)$. By integrating over all of the unknown parameters $\Theta$ we have:

$$p(D|B_S^h) = \int_\Theta p(\Theta|B_S^h)p(D|\Theta, B_S^h) \quad (4)$$

Researchers typically make a number of simplifying assumptions that collectively allow Equation 4 to be expressed in closed form. Before introducing these assumptions, we need the following notation.

As we showed in Section 2, if the local structure for a node $x_i$ is a decision graph $\mathcal{D}_i$, then sets of parameters $\Theta_{ij}$ and $\Theta_{ij'}$ can be identical for $j \neq j'$. For the derivations to follow, we find it useful to enumerate the *distinct* parameter sets in $\Theta_i$. Equivalently, we find it useful to enumerate the leaves in a decision graph.

For the remainder of this section, we adopt the following syntactic convention. When referring to a parameter set stored in the leaf of a decision graph, we use $a$ to denote the node index, and $b$ to denote the parent-state index. When referring to a parameter set in the context of a specific parent state of a node, we use $i$ to denote the node index and $j$ to denote the parent-state index.

To enumerate the set of leaves in a decision graph $\mathcal{D}_a$, we define a set of *leaf-set indices* $L_a$. The idea is that $L_a$ contains exactly one parent-state index for each leaf in the graph. More precisely, let $l$ denote the number of leaves in $\mathcal{D}_a$. Then $L_a = \{b_1, \ldots, b_l\}$ is defined as a set with the following properties:

1. For all $\{b, b'\} \subseteq L_a$, $b \neq b' \Rightarrow \Theta_{a,b} \neq \Theta_{a,b'}$
2. $\cup_{b \in L_a} \Theta_{a,b} = \Theta_a$

The first property ensures that each index in $L$ corresponds to a different leaf, and the second property ensures that every leaf is included.

One assumption used to derive Equation 4 in closed form is the *parameter independence* assumption. Simply stated, this assumption says that given the hypothesis $B_S^h$, knowledge about any distinct parameter set $\Theta_{ab}$ does not give us any information about any other distinct parameter set.

**Assumption 1 (Parameter Independence)**

$$p(\Theta|B_S^h) = \prod_{a=1}^{n} \prod_{b \in L_a} p(\Theta_{ab}|B_S^h)$$

Another assumption that researchers make is the *Dirichlet* assumption. This assumption restricts the prior distributions over the distinct parameter sets to be Dirichlet.

**Assumption 2 (Dirichlet)**
For all $a$ and for all $b \in L_a$,

$$p(\Theta_{ab}|B_S^h) \propto \prod_{c=1}^{r_a} \Theta_{abc}^{\alpha_{abc}-1}$$

where $\alpha_{abc} > 0$ for $1 \leq c \leq r_a$

Recall that $r_a$ denotes the number of states for node $x_a$. The hyperparameters $\alpha_{abc}$ characterize our prior knowledge about the parameters in $\Theta$. Heckerman et al. (1995) describe how to derive these exponents from a prior Bayesian network. We return to this issue later.

Using these assumptions, we can derive the Bayesian score for a structure that contains decision graphs by following a completely analogous method as Heckerman et al. (1995). Before showing the result, we must define the *inverse* function of $\Theta(i,j,k)$. Let $\theta$ denote an arbitrary parameter in $\Theta$. The function $\Theta^{-1}(\theta)$ denotes the set of index triples that $\Theta()$ maps into $\theta$. That is,

$$\Theta^{-1}(\theta) = \{i,j,k | \Theta(i,j,k) = \theta\}$$

Let $D_{ijk}$ denote the number of cases in $D$ for which $x_i = k$ and $Par(x_i) = j$. We define $N_{abc}$ as follows:

$$N_{abc} = \sum_{ijk \in \Theta^{-1}(\theta_{abc})} D_{ijk}$$

Intuitively, $N_{abc}$ is the number of cases in $D$ that provide information about the parameter $\theta_{abc}$. Letting $N_{ab} = \sum_c N_{abc}$ and $\alpha_{ab} = \sum_c \alpha_{abc}$, we can write the Bayesian score as follows:

$$p(D, B_S^h) = p(B_S^h) \prod_{a=1}^n \prod_{b \in L_a} \frac{\Gamma(\alpha_{ab})}{\Gamma(N_{ab} + \alpha_{ab})}$$
$$\cdot \prod_{c=1}^{|r_a|} \frac{\Gamma(N_{abc} + \alpha_{abc})}{\Gamma(\alpha_{abc})} \quad (5)$$

We can determine all of the counts $N_{abc}$ for each node $x_a$ as follows. First, initialize all the counts $N_{abc}$ to zero. Then, for each case $C$ in the database, let $k_C$ and $j_C$ denote the value for $x_i$ and $Par(x_i)$ in the case, respectively, and increment by one the count $N_{abc}$ corresponding to the parameter $\theta_{abc} = p(x_i = k_C | Par(x_i) = j_C, \Theta, \mathcal{D}_a)$. Each such parameter can be found efficiently by traversing $\mathcal{D}_a$ from the root.

We say a scoring function is *node decomposable* if it can be factored into a product of functions that depend only a node and its parents. Node decomposability is useful for efficiently searching through the space of global-network structures. Note that Equation 5 is node decomposable as long as $p(B_S^h)$ is node decomposable.

We now consider some node-decomposable distributions for $p(B_S^h)$. Perhaps the simplest distribution is to assume a uniform prior over network structures. That is, we set $p(B_S^h)$ to a constant in Equation 5. We use this simple prior for the experiments described in Section 5. Another approach is to (a-priori) favor networks with fewer parameters. For example, we can use

$$p(B_S^h) \propto \kappa^{|\Theta|} = \prod_{a=1}^n \kappa^{|\Theta_a|} \quad (6)$$

where $0 < \kappa <= 1$. Note that $\kappa = 1$ corresponds to the uniform prior over all structure hypotheses.

A simple prior for the parameters in $\Theta$ is to assume $\alpha_{abc} = 1$ for all $a, b, c$. This choice of values corresponds to a uniform prior over the parameters, and was explored by Cooper and Herskovits (1992) in the context of Bayesian networks containing complete tables. We call the Bayesian scoring function the *uniform scoring function* if all the hyperparameters are set to one. We have found that this prior works well in practice and is easy to implement.

Using two additional assumptions, Heckerman et al. (1995) show that each $\alpha_{abc}$ can be derived from a *prior Bayesian network*. The idea is that $\alpha_{abc}$ is proportional to the prior probability, obtained from the prior network, of all states of $\{x_i = k, Par(x_i) = j\}$ that map to the parameter $\theta_{abc}$. Specifically, if $B^P$ is our prior Bayesian network, we set

$$\alpha_{abc} = \alpha \sum_{ijk \in \Theta^{-1}(\theta_{abc})} p(x_i = k, Par(x_i) = j | B^P)$$

where $\alpha$ is a single *equivalent sample size* used to asses all of the exponents, and $Par(x_i)$ denotes the parents of $x_i$ in $\mathcal{G}$ (as opposed to the parents in the prior network). $\alpha$ can be understood as a measure of confidence that we have for the parameters in $B^P$. We call the Bayesian scoring function the *PN scoring function* (*P*rior *N*etwork scoring function) if the exponents are assessed this way. Heckerman et al. (1995) derive these constraints in the context of Bayesian networks with complete tables. In the full version of this paper, we show that these constraints follow when using decision graphs as well, with only slight modifications to the additional assumptions.

Although we do not provide the details here, we can use the decision-graph structure to efficiently compute the exponents $\alpha_{abc}$ from the prior network in much the same way we computed the $N_{abc}$ values from the database.

## 4 SEARCH

Given a scoring function that evaluates the merit of a Bayesian-network structure $B_S$, learning Bayesian networks from data reduces to a search for one or more structures that have a high score. Chickering (1995) shows that finding the optimal structure containing complete tables for the mappings $M$ is NP-hard when using a Bayesian scoring function. Given this result, it seems reasonable to assume that by allowing (the more general) decision-graph mappings, the problem remains hard, and consequently it is appropriate to apply heuristic search techniques.

In Section 4.1, we define a search space over decision-graph structures within a single node $x_i$, assuming that the parent set $Par(x_i)$ is fixed. Once such a space is defined, we can apply to that space any number of well-known search algorithms. For the experiments described in Section 5, for example, we apply greedy search.

In Section 4.2 we describe a greedy algorithm that combines local-structure search over all the decision graphs in the nodes with a global-structure search over the edges in $\mathcal{G}$.

## 4.1 DECISION-GRAPH SEARCH

In this section, we assume that the states of our search space correspond to all of the possible decision graphs for some node $x_i$. In order for a search algorithm to traverse this space, we must define a set of operators that transform one state into another.

There are three operators we define, and each operator is a modification to the current set of leaves in a decision graph.

**Definition (Complete Split)**
*Let $v$ be a leaf node in the decision graph, and let $\pi \in Par(x_i)$ be a parent of $x_i$. A complete split $C(v, \pi)$ adds $r_i$ new leaf nodes as children to $v$, where each child of $v$ corresponds to a distinct value of $\pi$.*

**Definition (Binary Split)**
*Let $v$ be a leaf node in the decision graph, and let $\pi \in Par(x_i)$ be a parent of $x_i$. A binary split $B(v, \pi, k)$ adds 2 new leaf nodes as children to $v$, where the first child corresponds to state $k$ of $\pi$, and the other child corresponds to all other states of $\pi$.*

**Definition (Merge)**
*Let $v_1$ and $v_2$ be two distinct leaf nodes in the decision graph. A Merge $M(v_1, v_2)$ merges the $v_1$ and $v_2$ into a single node. That is, the resulting node inherits all parents from both $v_1$ and $v_2$.*

In Figure 4, we show the result of each type of operator to a decision graph for a node $z$ with parents $x$ and $y$, where $x$ and $y$ both have three states.

We add the pre-condition that the operator must change the parameter constraints implied by the decision graph. We would not allow, for example, a complete split $C(v_1, y)$ in Figure 4a: two of $v_1$'s new children would correspond to impossible states of $y$ ($\{y = 0$ and $y = 1\}$ and $\{y = 0$ and $y = 2\}$), and the third child would correspond to the original constraints at $v_1$ ($\{y = 0$ and $y = 0\}$).

Note that starting from a decision graph containing a single node (both the root and a leaf node), we can generate a complete decision tree by repeatedly applying complete splits. As discussed in the previous section, we can represent any parameter-set equalities by merging the leaves of a complete decision tree. Consequently, starting from a graph containing one node there exists a series of operators that result in any set of possible parameter-set equalities. Note also that if we repeatedly merge the leaves of a decision graph until there is a single parameter set, the resulting graph is equivalent (in terms of parameter equalities) to the graph containing a single node. Therefore, our operators are sufficient for moving from any set of parameter constraints to any other set of parameter constraints. Although we do not discuss them here, there are methods that can simplify (in terms of the number of nodes) some decision graphs such that they represent the same set of parameter constraints.

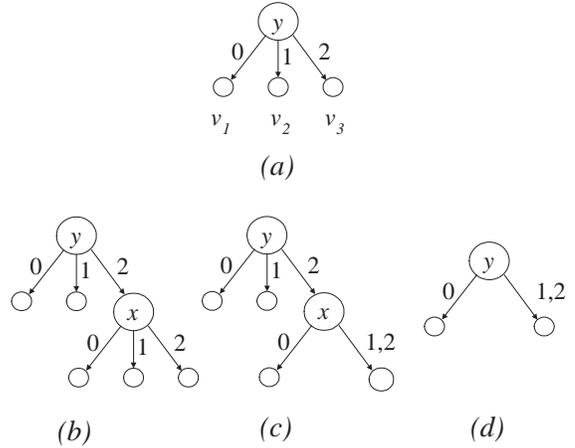

Figure 4: Example of the application of each type of operator: (a) the original decision graph, (b) the result of applying $C(v_3, x)$, (c) the result of applying $B(v_3, x, 0)$, and (d) the result of applying $M(v_2, v_3)$

The complete-split operator is actually not needed to ensure that all parameter equalities can be reached: any complete split can be replaced by a series of binary splits such that the resulting parameter-set constraints are identical. We included the complete-split operator in the hopes that it would help lead the search algorithm to better structures. In Section 5, we compare greedy search performance in various search spaces defined by including only subsets of the above operators.

## 4.2 COMBINING GLOBAL AND LOCAL SEARCH

In this section we describe a greedy algorithm that combines global-structure search over the edges in $\mathcal{G}$

with local-structure search over the decision graphs in all of the nodes of $\mathcal{G}$.

Suppose that in the decision-graph $\mathcal{D}_i$ for node $x_i$, there is no non-leaf node annotated with some parent $\pi \in Par(x_i)$. In this case, $x_i$ is independent of $\pi$ given its other parents, and we can remove $\pi$ from $Par(x_i)$ without violating the decomposition given in Equation 1. Thus given a fixed structure, we can learn all the local decision graphs for all of the nodes, and then delete those parents that are independent. We can also consider adding edges as follows. For each node $x_i$, add to $Par(x_i)$ all non-descendants of $x_i$ in $\mathcal{G}$, learn a decision graph for $x_i$, and then delete all parents that are not contained in the decision graph. Figure 5 shows a greedy algorithm that uses combines these two ideas. In our experiments, we started the algorithm with a structure for which $\mathcal{G}$ contains no edges, and each graph $\mathcal{D}_i$ consists of a single root node.

1. Score the current network structure $B_S$
2. For each node $x_i$ in $\mathcal{G}$
3.     Add every non-descendant that is not a parent of $x_i$ to $Par(x_i)$
4.     For every possible operator $O$ to the decision graph $\mathcal{D}_i$
5.         Apply $O$ to $B_S$
6.         Score the resulting structure
7.         Unapply $O$
8.     Remove any parent that was added to $x_i$ in step 3
9.     If the best score from step 6 is better than the current score
10.         Let $O$ be the operator that resulted in the best score
11.         If $O$ is a split operator (either complete or binary) on a node $x_j$ that is not in $Par(x_i)$, then add $x_j$ to $Par(x_i)$
12.         Apply $O$ to $B_S$
13.         Goto 1
14.     Otherwise, return $B_S$

Figure 5: Greedy algorithm that combines local and global structure search

Note that as a result of a merge operator in a decision graph $\mathcal{D}_i$, $x_i$ may be rendered independent from one of its parents $\pi \in Par(x_i)$, even if $\mathcal{D}_i$ contains a node annotated with $\pi$. For a simple example, we could repeatedly merge all leaves into a single leaf node, and the resulting graph implies that $x_i$ does not depend on any of its parents. We found experimentally that—when using the algorithm from Figure 5—this phenomenon is rare. Because testing for these parent deletions is expensive, we chose to not check for them in the experiments described in Section 5.

Another greedy approach for learning structures containing decision trees has been explored by Friedman and Goldszmidt (1996). The idea is to score edge operations in $\mathcal{G}$ (adding, deleting, or reversing edges) by applying the operation and then greedily learning the local decision trees for any nodes who's parents have changed as a result of the operation. In the full version of the paper, we compare our approach to theirs.

## 5 EXPERIMENTAL RESULTS

In this section we investigate how varying the set of allowed operators affects the performance of greedy search. By disallowing the merge operator, the search algorithms will identify decision-tree local structures in the Bayesian network. Consequently, we can see how learning accuracy changes, in the context of greedy search, when we generalize the local structures from decision trees to decision graphs.

In all of the experiments described in this section, we measure learning accuracy by the posterior probability of the identified structure hypotheses. Researchers often use other criteria, such as predictive accuracy on a holdout set or structural difference from some generative model. The reason that we do not use any of these criteria is that we are evaluating *how well the search algorithm performs in various search spaces*, and the goal of the search algorithm is to *maximize the scoring function*. We *are not* evaluating how well the Bayesian scoring functions approximate some other criteria.

In our first experiment, we consider the *Promoter Gene Sequences* database from the UC Irvine collection, consisting of 106 cases. There are 58 variables in this domain. 57 of these variables, $\{x_1, \ldots, x_{57}\}$ represent the "base-pair" values in a DNA sequence, and each has four possible values. The other variable, *promoter*, is binary and indicates whether or not the sequence has promoter activity. The goal of learning in this domain is to build an accurate model of the distribution $p(promoter|x_1, \ldots, x_{57})$, and consequently it is reasonable to consider a static graphical structure for which $Par(promoter) = \{x_1, \ldots, x_{57}\}$, and search for a decision graph in node *promoter*.

Table 1 shows the relative Bayesian scores for the best decision graph learned, using a greedy search with various parameter priors and search spaces. All searches started with a decision graph containing a single node, and the current best operator was applied at each step until no operator increased the score of the current state. Each column corresponds to a different restriction of the search space described in Section 4.1: the labels indicate what operators the greedy search was allowed to use, where C denotes complete splits, B denotes binary splits, and M denotes merges. The column labeled BM, for example, shows the results when a greedy search used binary splits and merges, but not complete splits. Each row corresponds to a different parameter-prior for the Bayesian scoring function. The U-PN scoring function is a special case of the PN scoring function for which the prior network imposes a uniform distribution over all variables. The number following the U-PN in the row labels indicates the equivalent-sample size $\alpha$. All results use a uniform prior over structure hypotheses. A value of zero in a row of the table denotes the hypothesis with lowest probability out of all those identified using the given parameter prior. All other values denote the natural logarithm of how many times more likely the identified hypothesis is than the one with lowest probability.

Table 1: Greedy search performance for various Bayesian scoring functions, using different sets of operators, in the *Promoter* domain.

|         | C | B     | CB   | CM    | BM    | CBM   |
|---------|---|-------|------|-------|-------|-------|
| uniform | 0 | 13.62 | 6.07 | 22.13 | 26.11 | 26.11 |
| U-PN 10 | 0 | 6.12  | 4.21 | 9.5   | 10.82 | 12.93 |
| U-PN 20 | 0 | 5.09  | 3.34 | 14.11 | 12.11 | 14.12 |
| U-PN 30 | 0 | 4.62  | 2.97 | 10.93 | 12.98 | 16.65 |
| U-PN 40 | 0 | 3.14  | 1.27 | 16.3  | 13.54 | 16.02 |
| U-PN 40 | 0 | 2.99  | 1.12 | 15.76 | 15.54 | 17.54 |

By comparing the relative values between searches that use merges and searches that *don't* use merges, we see that without exception, adding the merge operator results in a significantly more probable structure hypothesis. We can therefore conclude that a greedy search over decision graphs results in better solutions than a greedy search over decision trees. An interesting observation is that complete-split operator actually reduces solution quality when we restrict the search to decision trees.

We performed an identical experiment to another classification problem, but for simplicity we only present the results for the uniform scoring function. Recall from Section 3 that the uniform scoring function has all of the hyperparameters $\alpha_{abc}$ set to one. This second experiment was run with the *Splice-junction Gene Sequences* database, again from the UC Irvine repository. This database also contains a DNA sequence, and the problem is to predict whether the position in the middle of the sequence is an "intron-exon" boundary, an "exon-intron" boundary, or neither. The results are given in Table 2. We used the same uniform prior for structure hypotheses.

Table 2: Greedy search performance for the uniform scoring function, using different sets of operators, in the *Splice* domain.

| C | B   | CB  | CM  | BM  | CBM |
|---|-----|-----|-----|-----|-----|
| 0 | 383 | 363 | 464 | 655 | 687 |

Table 3: Greedy search performance for the uniform scoring function for each node in the ALARM network. Also included is the uniform score for the complete-table model

| COMP | C   | B   | CB  | CM  | BM  | CBM |
|------|-----|-----|-----|-----|-----|-----|
| 0    | 134 | 186 | 165 | 257 | 270 | 270 |

Table 2 again supports the claim that we get a significant improvement by using decision graphs instead of decision trees.

Our final set of experiments were done in the ALARM domain, a well-known benchmark for Bayesian-network learning algorithms. The ALARM network, described by Beinlich et al. (1989), is a hand-constructed Bayesian network used for diagnosis in a medical domain. The parameters of this network are stored using complete tables.

In the first experiment for the ALARM domain, we demonstrate that for a fixed global structure $\mathcal{G}$, the hypothesis identified by searching for local decision graphs in all the nodes can be significantly better than the hypothesis corresponding to complete tables in the nodes. We first generated 1000 cases from the ALARM network, and then computed the uniform Bayesian score for the ALARM network, assuming that the parameter mappings $M$ are complete tables. We expect the posterior of this model to be quite good, because we're evaluating the *generative* model structure. Next, using the uniform scoring function, we applied the six greedy searches as in the previous experiments to identify good decision graphs for *all* of the nodes in the network. We kept the global structure $\mathcal{G}$ fixed to be identical to the global structure of the ALARM network. The results are shown in Table 3, and the values have the same semantics as in the previous two tables. The score given in the first column labeled COMP is the score for the complete-table model.

Table 3 demonstrates that search performance using decision graphs can identify significantly better models than when just using decision trees. The fact that the complete-table model attains such a low score (the best hypothesis we found is $e^{270}$ times more probable than the complete-table hypothesis!) is not surprising upon examination of the probability tables stored

Table 4: Performance of greedy algorithm that combines local and global structure search, using different sets of operators, in the ALARM domain. Also included is the result of a greedy algorithm that searches for global structure assuming complete tables.

| COMP | C | B | CB | CM | BM | CBM |
|---|---|---|---|---|---|---|
| 255 | 0 | 256 | 241 | 869 | 977 | 1136 |

Table 5: Performance of a restricted version of our greedy algorithm, using different sets of operators, in the ALARM domain. Also included is the result of a greedy algorithm, initialized with the global structure of the ALARM network, that searches for global structure assuming complete tables.

| COMP | C | B | CB | CM | BM | CBM |
|---|---|---|---|---|---|---|
| 0 | 179 | 334 | 307 | 553 | 728 | 790 |

in the ALARM network: most of the tables contain parameter-set equalities.

In the next experiment, we used the ALARM domain to test the structure-learning algorithm given in Section 4.2. We again generated a database of 1000 cases, and used the uniform scoring function with a uniform prior over structure hypotheses. We ran six versions of our algorithm, corresponding to the six possible sets of local-structure operators as in the previous experiments. We also ran a greedy structure-search algorithm that assumes complete tables in the nodes. We initialized this search with a global network structure with no edges, and the operators were single-edge modifications to the graph: deletion, addition and reversal. In Table 4 we show the results. The column labeled COMP corresponds to the greedy search over structures with complete tables.

Once again, we note that when we allow nodes to contain decision graphs, we get a significant improvement in solution quality. Note that the search over complete-table structures out-performed our algorithm when we restricted the algorithm to search for decision trees containing either (1) only complete splits or (2) complete splits and binary splits.

In our final experiment, we repeated the previous experiment, except that we only allowed our algorithm to add parents that are not descendants in the generative model. That is, we restricted the global search over $\mathcal{G}$ to those dags that did not violate the partial ordering in the ALARM network. We also ran the same greedy structure-search algorithm that searches over structures with complete tables, except we initialized the search with the ALARM network. The results of this experiment are shown in Table 5. From the table, we see that the constrained searches exhibit the same relative behavior as the unconstrained searches.

For each experiment in the ALARM domain (Tables 3, 4, and 5) the values presented measure the performance of search relative to the worst performance in that experiment. In Table 6, we compare search performance across all experiments in the ALARM domain. That is, a value of zero in the table corresponds to the experiment and set of operators that led to the learned hypothesis with lowest posterior probability, out of all experiments and operator restrictions we considered in the ALARM domain. All other values given in the table are relative to this (lowest) posterior probability. The row labels correspond to the experiment: S denotes the first experiment that performed local searches in a static global structure, U denotes the second experiment that performed unconstrained structural searches, and C denotes the final experiment that performed constrained structural search.

Table 6: Comparison of Bayesian scores for all experiments in the ALARM domain

|   | COMP | C | B | CB | CM | BM | CBM |
|---|---|---|---|---|---|---|---|
| S | 278 | 412 | 464 | 443 | 534 | 548 | 548 |
| U | 255 | 0 | 256 | 241 | 869 | 976 | 1136 |
| C | 336 | 515 | 670 | 643 | 889 | 1064 | 1126 |

Rather surprising, each hypothesis learned using global-structure search with decision graphs had a higher posterior than every hypothesis learned using the *generative* static structures.

## 6 DISCUSSION

In this paper we showed how to derive the Bayesian score of a network structure that contains parameter maps implemented as decision graphs. We defined a search space for learning individual decision graphs within a static global structure, and defined a greedy algorithm that searches for both global and local structure simultaneously. We demonstrated experimentally that greedy search over structures containing decision graphs significantly outperforms greedy search over both (1) structures containing complete tables and (2) structures containing decision trees.

We now consider an extension to the decision graph that we mentioned in Section 2.3. Recall that in a decision graph, the parameter sets are stored in a table within the leaves. When decision graphs are implemented this way, any parameter $\theta_{abc}$ must belong to exactly one (distinct) parameter set. An important consequence of this property is that if the priors for the parameter sets are Dirichlet (Assumption 2), then the posterior distributions are Dirichlet as well. That is, the Dirichlet distribution is *conjugate* with respect to the likelihood of the observed data. As a result, it is easy to derive the Bayesian scoring function in closed form.

If we allow nodes within a decision graph $\mathcal{D}_i$ to split on node $x_i$, we can represent an arbitrary set of parameter constraints of the form $\Theta(i,j,k) = \Theta(i,j',k')$ for $j \neq j'$ and $k \neq k'$. For example, consider a Baysian network for the two-variable domain $\{x,y\}$, where $x$ is a parent of $y$. We can use a decision graph for $y$ that splits on $y$ to represent the constraint

$$p(y=1|x=0,\Theta,\mathcal{D}_y,\mathcal{G}) = p(y=0|x=1,\Theta,\mathcal{D}_y,\mathcal{G})$$

Unfortunately, when we allow these types of constraints, the Dirichlet distribution is no longer conjugate with respect to the likelihood of the data, and the parameter independence assumption is violated. Consequently, the derivation described in Section 3 will not apply. Conjugate priors for a decision graph $\mathcal{D}_i$ that splits on node $x_i$ do exist, however, and in the full version of this paper we use a weaker version of parameter independence to derive the Bayesian score for these graphs in closed form.

We conclude by noting that it is easy to extend the definition of a network structure to represent constraints between the parameters of different nodes in the network, e.g. $\Theta_{ij} = \Theta_{i'j'}$ for $i \neq i'$. Both Buntine (1994) and Thiesson (1995) consider these types of constraints. The Bayesian score for such structures can be derived by simple modifications to the approach described in this paper.